\documentclass[runningheads]{llncs}

\usepackage{algorithm}  
\usepackage{algpseudocode}
\usepackage{makecell}
\usepackage{multirow}
\usepackage{caption}
\usepackage{tabularx}
\usepackage{accv}

\usepackage{accvabbrv}

\usepackage{booktabs}

\usepackage[accsupp]{axessibility}  

\usepackage[pagebackref,breaklinks,colorlinks,citecolor=accvblue]{hyperref}

\usepackage{orcidlink}

\newcommand{\figref}[1]{Fig. \ref{#1}}
\newcommand{\tableref}[1]{Table \ref{#1}}
\newcommand{\algoref}[1]{Algorithm \ref{#1}}

\begin{document}

\title{Parameter-Selective Continual\\Test-Time Adaptation} 



\author{Jiaxu Tian\inst{1} \and
Fan Lyu\inst{2}\thanks{Corresponding author.}
\authorrunning{J. Tian et al.}
\institute{College of Electric and Information, Northeast Agricultural University\and
New Laboratory of Pattern Recognition, Institute of Automation, Chinese Academy of Sciences\\
\email{jiaxutian@neau.edu.cn} \quad \email{fan.lyu@cripac.ia.ac.cn}}
}

\maketitle
\begin{abstract}
 Continual Test-Time Adaptation (CTTA) aims to adapt a pretrained model to ever-changing environments during the test time under continuous domain shifts.
Most existing CTTA approaches are based on the Mean Teacher (MT) structure, which contains a student and a teacher model, where the student is updated using the pseudo-labels from the teacher model, and the teacher is then updated by exponential moving average strategy. 
However, these methods update the MT model indiscriminately on all parameters of the model.
That is, some critical parameters involving sharing knowledge across different domains may be erased, intensifying error accumulation and catastrophic forgetting. 
In this paper, we introduce Parameter-Selective Mean Teacher (PSMT) method, which is capable of effectively updating the critical parameters within the MT network under domain shifts.
First, we introduce a selective distillation mechanism in the student model, which utilizes past knowledge to regularize novel knowledge, thereby mitigating the impact of error accumulation. 
Second, to avoid catastrophic forgetting, in the teacher model, we create a mask through Fisher information to selectively update parameters via exponential moving average, with preservation measures applied to crucial parameters. 
Extensive experimental results verify that PSMT outperforms state-of-the-art methods across multiple benchmark datasets. Our code is available at \url{https://github.com/JiaxuTian/PSMT}.
\end{abstract}

\section{Introduction}
\label{sec:intro}
Continual Test-Time Adaptation (CTTA) \cite{CoTTA,lyu2024variational} is a novel task that involves adapting a pretrained source model during the testing time under continuous domain shifts.
For any deployed model, CTTA is valuable since it maintains the model in dynamic real-world environments where data distributions can vary significantly over time. 
CTTA has been studied in many applications, such as surveillance systems~\cite{surveillance}, automated driving systems~\cite{driving,driving2}, medical image segmentation~\cite{medical1} and natural catastrophes~\cite{nature}.

\begin{figure}[t]
	\includegraphics[width=\textwidth]{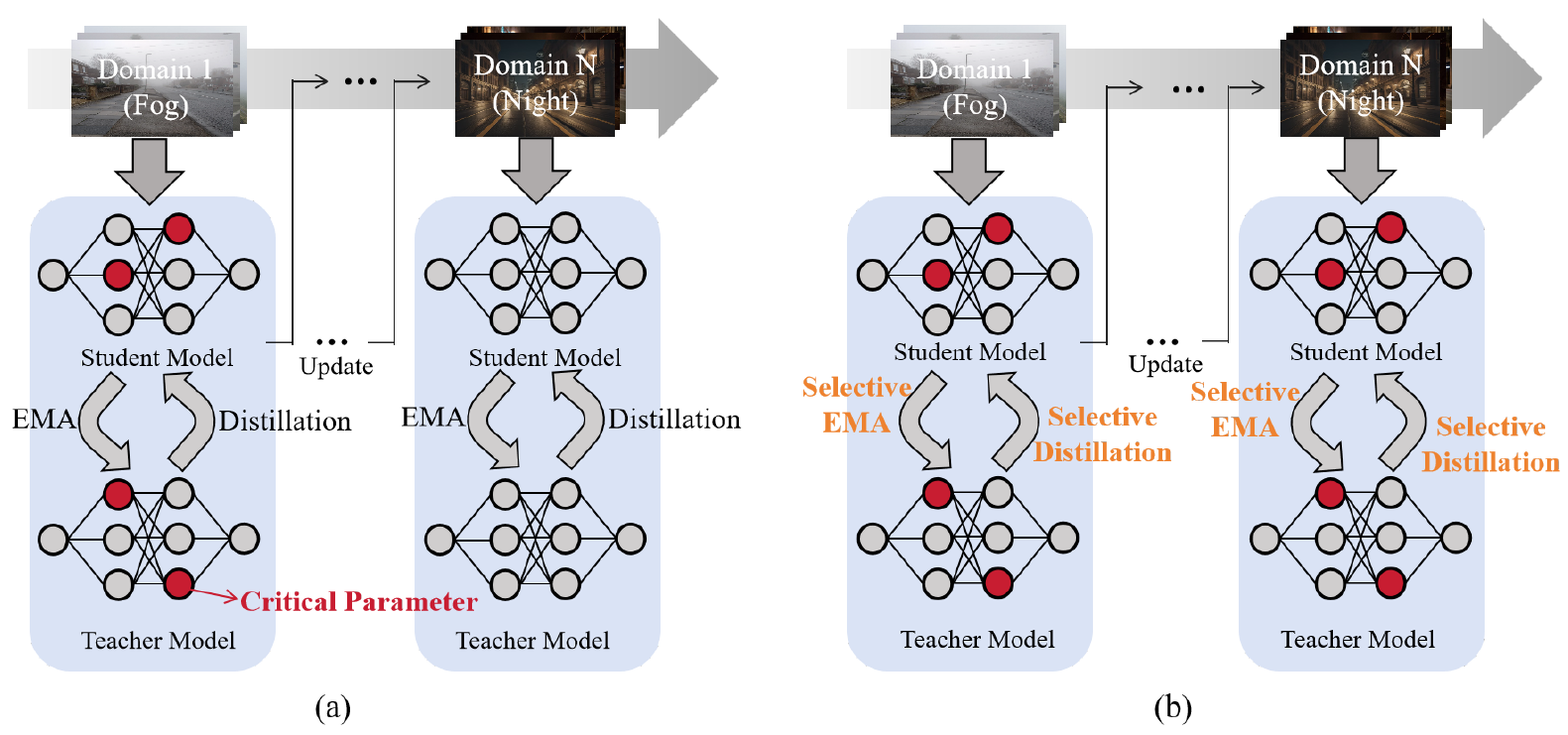}
	\caption{\textbf{(a)} In the traditional MT-based methods, all parameters are updated, which can be problematic for CTTA tasks. This approach may result in the phenomenon of error accumulation. \textbf{(b)} Our method improves on the issue of updating crucial parameters by selectively restoring them instead. The student model focuses on acquiring new knowledge efficiently, while the teacher model is dedicated to reinforcing previous knowledge. } \label{problem}
\end{figure}

To enable a deployed model to adapt under unsupervised domain changes, a well-proved architecture, i.e. mean teacher (MT) \cite{mean}, is used.
MT methods~\cite{CoTTA,rmt,DSS,tri,PETAL,ess,shi2024controllable} build a teacher model to provide pseudo labels for a student model of the same structure.
The student model learns from these pseudo labels, improving its predictions towards these provided labels. 
For instance, CoTTA \cite{CoTTA} employs weight-averaged and augmentation-averaged predictions to address challenges of error accumulation \cite{error} and catastrophic forgetting \cite{catastrophic,catastrophic2,catastrophic3}, effectively maintaining essential foundational knowledge and adapting to domain shifts \cite{domainshift,domainshift1}. 
RMT \cite{rmt} method adopts symmetric cross-entropy loss \cite{sce} over traditional cross-entropy loss \cite{ce} for self-training \cite{selftrain,selftrain1}, further enhancing robustness against frequent and varied domain shifts. 

Although existing methods based on MT approach excel in CTTA tasks, they always update the teacher and the student model undergoes the whole parameters for adaptation (as illustrated in \figref{problem}(a)).
Liu et al. \cite{overparam} indicate that in over-parameterized neural networks, only a subset of parameters is truly effective, while the remainder may even deteriorate model performance.
Inspired by this, in CTTA tasks, we assume crucial parameters exist in the neural network that are crucial for representing the shared information across different tasks, and these parameters are only a small part of the whole model.
Updating crucial parameters indiscriminately may lead to the overwriting of overlapping knowledge from previous and current tasks, which results in catastrophic forgetting~\cite{catastrophic,catastrophic2,lyu2021multi,lyu2023measuring} and the accumulation of errors~\cite{error}.
Thus, parameters may misalign the model’s ability that perform well on original tasks, ultimately diminishing the model’s effectiveness in CTTA tasks.

To tackle the above issue, 
we propose a novel approach termed Parameter-Selective Mean Teacher (PSMT) under the crucial parameter assumption. 
As shown in Fig.~\ref{problem}(b), guided by the objective of retaining crucial parameters, in the MT structure, we select the Fisher information for identifying important parameters. 
Specifically, the core concept of PSMT revolves around the dynamic update of both the teacher and student models.
First, the student model dynamically calculates the importance of model parameters in response to new data, applying regularization to maintain previously learned tasks, thus reducing the risk of overfitting \cite{overfit}.
Second, the teacher model focuses on parameter restoration by selectively preserving crucial parameters during learning progress, which helps in stabilizing past knowledge and reducing catastrophic forgetting. Overall, PSMT balances the need for model adaptation to new information while preserving the integrity of previously acquired knowledge, thereby effectively mitigating the issues of catastrophic forgetting and error accumulation. Experimental results clearly demonstrate that PSMT outperforms existing state-of-the-art methods. 

In summary, our main contributions of this paper are as follows:
\begin{itemize}
	\item We propose a Parameter-Selective Mean Teacher (PSMT) method that dynamically selects crucial parameters for effective update in the test-time adaptation.
	\item We introduce a parameter-selective mechanism to update the student model, which utilizes this mechanism to regularize past and existing knowledge, thereby effectively preventing overfitting of current knowledge.
	\item We introduce a parameter-selective mechanism to update the teacher model, a mechanism that quantifies the importance of parameters to retain the crucial ones selectively. This mechanism enhances stability when learning new tasks and ensures the preservation of previously acquired knowledge.
\end{itemize}

\section{Related Work}

\subsection{Continual Test-Time Adaptation}
The goal of CTTA \cite{CoTTA,tan2024less} is to continuously adapt a pretrained source model to changing target domains, utilizing unlabeled data from these new environments, thereby mitigating performance degradation due to domain shifts \cite{domainshift,domainshift1}.
CoTTA \cite{CoTTA} emerged as a pioneering effort to tackle the problem of CTTA, utilizing the MT approach \cite{mean}, CoTTA improves the quality of pseudo-labels \cite{pseudo} with weight-averaged predictions and employs a stochastic parameter reset mechanism, selectively restoring model parameters to mitigate knowledge loss during significant domain shifts. Building on CoTTA's groundwork, PETAL \cite{PETAL} proposed a comprehensive probabilistic framework. Distinguishing itself with a data-driven parameter restoration technique \cite{datadriven} based on the Fisher information matrix (FIM), PETAL efficiently combats catastrophic forgetting. Wang et al. \cite{DSS} introduced a novel Dynamic Sample Selection method for CTTA that effectively mitigates error accumulation by employing dynamic thresholding to distinguish between high and low-quality samples. While existing methods based on the MT demonstrate commendable performance in CTTA tasks, the inherent mechanism of updating crucial parameters throughout successive iterations invariably leads to error accumulation \cite{error} and catastrophic forgetting \cite{catastrophic,catastrophic2,catastrophic3}.

\subsection{Parameter Selection in Neural Network}
In neural networks, the phenomenon of over-parameterization \cite{overparam} is often observed, where only a subset of the parameters contributes to the network's performance. Consequently, numerous methods have been developed to address this issue. Among these methods, gradient accumulation \cite{grad} evaluates the significance of parameters based on the accumulated sum of the squared gradients during training, and sensitivity analysis \cite{sensitivity}, on the other hand, measures the effect of infinitesimal changes in parameters on the output, identifying parameters that significantly affect the network’s behavior.
However, gradient accumulation and sensitivity analysis need large computational costs and lack a robust theoretical basis that connects to optimization methods in neural network training \cite{neural}. 

Fisher Information (FI) is a concept in statistical estimation theory, where it measures the sensitivity of the likelihood function to changes in model parameters and quantifies how much information observed data provides about a parameter within a model. Moreover, FI is supported by theoretical underpinnings in statistical estimation theory.
For instance, Ly et al. \cite{ly} demonstrated how FI can be instrumental in pinpointing models that perform best on a given dataset, playing a crucial role in preventing both overfitting \cite{overfit} and underfitting \cite{underfitting}. Liu et al. \cite{liu} argued that FI can lead to improved outcomes by leveraging curvature information. This perspective underscores the FI's capability to provide a more nuanced understanding of the parameter space, potentially enhancing model optimization \cite{fisheroptim}. However, existing FI methods are mainly focused on scenarios with labels \cite{fisherlabel}, rarely applying them in situations without labels. Recently, Niu et al.~\cite{eata} proposed a weighted Fisher regularizer in the CTTA tasks utilizing FI. This regularizer prevents significant changes in parameters that are important for the in-distribution domain. However, it ignores the error accumulation issues. Instead, our PSMT studies to leverage FI in a MT architecture, which has been proven effectiveness to error accumulation.

\section{Method}
\subsection{Overview}
Given a model $f_{\theta_{0}}$, parameterized by $\theta_{0}$ and initially trained on a source dataset $\mathcal{D}^\mathcal{S}=(\mathcal{X}^\mathcal{S},\mathcal{Y}^\mathcal{S})$, we consider the unlabeled target domain $\mathcal{U}^\mathcal{T}=\{\mathcal{U}^{1},\mathcal{U}^{2},\cdots,\mathcal{U}^{i},\\\cdots,\mathcal{U}^{N}\}$, which consists of multiple domains. Our objective can be described as: at time step $t$, the model $f_{\theta_{t}}$ is tasked with adapting to the newly introduced,  the samples $x_{t}\in\mathcal{U}^\mathcal{T}$. Consequently, the model is expected not only to make predictions $f_{\theta_{t}}(x_{t})$ but also to undergo self-adaptation in preparation for subsequent datasets, transitioning from $\theta_{t}$ to $\theta_{t+1}$.

\begin{figure}[t]
	\includegraphics[width=\textwidth]{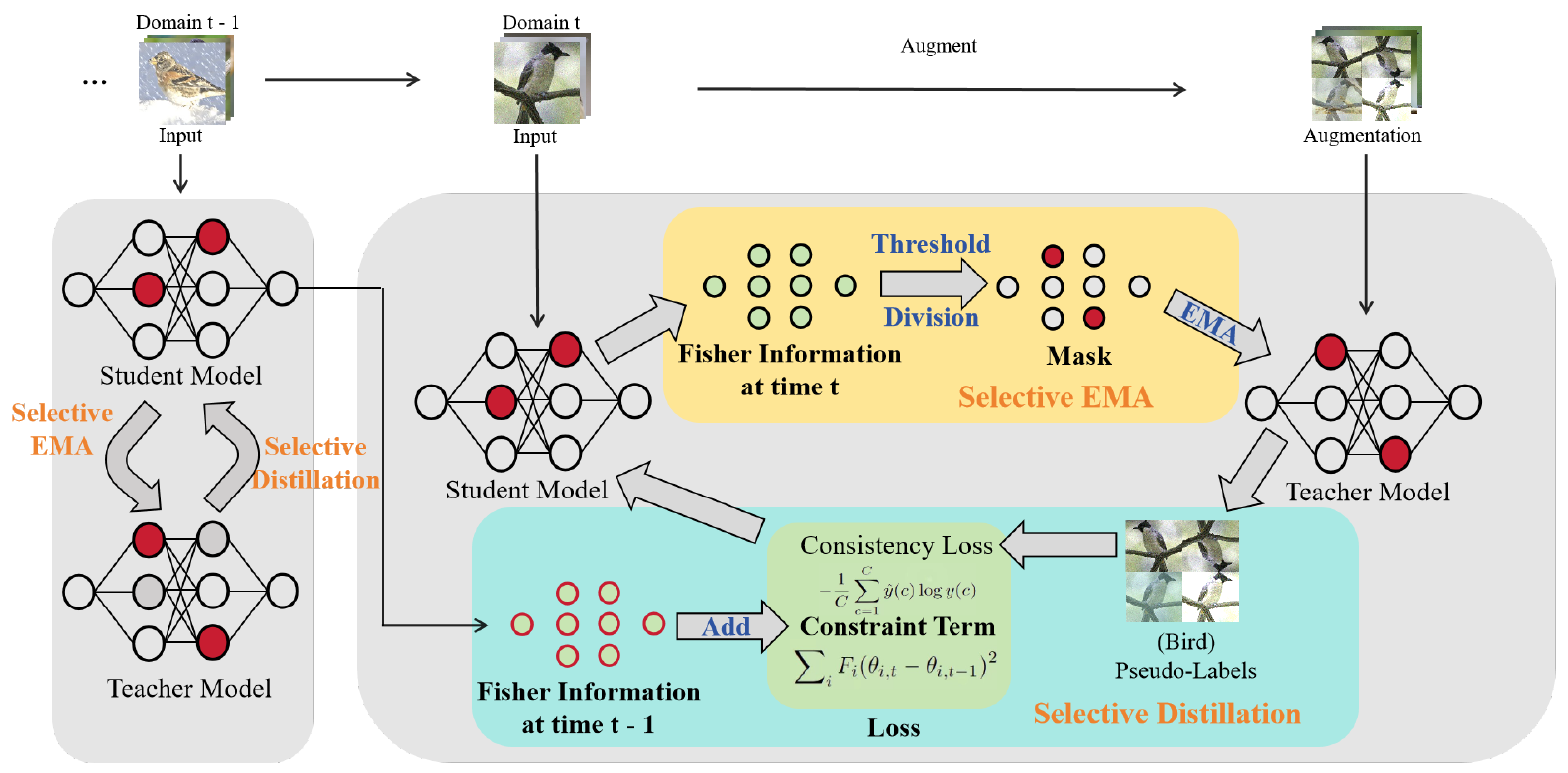}
	\caption{\textbf{PSMT framework: }Using test samples as inputs for the student model and augmented samples for the teacher model, PSMT enhances the conventional student model through regularization of existing knowledge using past knowledge. PSMT improves the traditional EMA method by selecting crucial parameters based on Fisher information.} \label{net}
\end{figure}

Prevalent methods for achieving this goal have predominantly employed MT structures. These models generally involve a structure where a teacher network guides the learning of a student network through consistency regularization, effectively smoothing the learning trajectory over time. 
However, these traditional MT methods may update crucial parameters, which can lead to error accumulation and catastrophic forgetting, significantly hampering the system's ability to preserve crucial parameters essential for task performance. 
To achieve the aforementioned goal and address this issue, we propose a PSMT method, a strategy that utilizes Fisher information differently in the student and teacher models. The student model leverages Fisher information to enhance the acquisition of new knowledge, focusing on efficiency in learning. Additionally, the teacher model employs Fisher information to strengthen and stabilize previously acquired knowledge, as depicted in \figref{net}.
In the following, we illustrate the details of  our PSMT.
\subsection{Student Update using Selective Distillation}
In CTTA, the conventional student model may update crucial parameters to adjust to changes in the domains. However, this mechanism could update crucial parameters, which could increase the buildup of errors and lead to catastrophic forgetting, where the model loses previously learned information.
To avoid incorrect updates of crucial parameters, we propose a mechanism to enable real-time evaluation of the importance of model parameters under current conditions and combine previously learned knowledge to regularize new knowledge. 

In the methodology for calculating parameter's importance, we select the FIM, which is frequently utilized as a criterion for assessing the significance of parameters. We employ the FIM as the method to evaluate the importance of model parameters at the time $t-1$. When utilizing FIM to assess the importance of model parameters, the elements on the diagonal play a crucial role. These diagonal elements indicate the sensitivity of the data to variations in each independent parameter, thereby reflecting the contribution of each parameter to the model's predictive power. Higher values on the diagonal suggest that a parameter is more significant within the model, as this implies that small changes in the parameter will lead to significant changes in the prediction outcomes. To compute the FIM, it is essential to first calculate the gradient of the log-likelihood with respect to the parameter vector $\theta_t$ for the predicted output $p_{\theta_{t-1}}(x_{t-1})$, where $x_{t-1}$ represents a batch of past test samples. The FIM is then formulated as follows:
\begin{equation}\label{fim}
    F = {\rm{Diag}}[\nabla\log p_{\theta_{t-1}}(x_{t-1})(\nabla\log p_{\theta_{t-1}}(x_{t-1}))^{T}].
\end{equation}

Traditional distillation is conducted by the teacher model through the generation of pseudo-labels, without imposing constraints on the learning of current knowledge. By utilizing the diagonal FIM, we can adopt certain methods to constrain the potential overfitting that may occur when learning new knowledge. Quadratic constraints are particularly beneficial because they provide localized and bounded adjustments to the parameters, maintaining continuous and differentiable gradients that facilitate a more stable and efficient learning process. 
Unlike linear constraints, which may not offer sufficient guidance for parameter adjustments to improve model performance, quadratic constraints possess desirable mathematical properties such as convexity. This ensures the possibility of finding global optima and simplifies computations. Therefore, making quadratic constraints a more robust choice for maintaining balance between preserving learned knowledge and accommodating new information. Simultaneously, inspired by~\cite{EWC}, our quadratic constraint term is formulated as follows:
\begin{equation}\label{stuloss}
    \mathcal{L}_{\text{stu}}=\sum\limits\nolimits_{i}F_{i}(\theta_{i,t}-\theta_{i,t-1})^2,
\end{equation}
where $F_i$ represents FIM associated with the 
$i^{th}$ parameter and $\theta_{i,t}$ is the current value of the $i^{th}$ parameter in the model at time $t$. Through this mechanism, we can enhance the performance of the student model in acquiring new knowledge, by avoiding updates to important parameters, thereby mitigating the impact of domain shift on the model.

\subsection{Teacher Update using Selective Exponential Moving Average}
The teacher model in the MT method also encounters the drawback of updating crucial parameters. It updates through the exponential moving average (EMA): 
\begin{equation}\label{ema}
    \theta_{t+1}^{'}=\delta\theta_{t}^{'}+(1-\delta)\theta_{t+1},
\end{equation}
where $\delta$ denotes a smoothing coefficient, $\theta_{t}^{'}$ corresponds to the teacher model at time $t$ and $\theta_{t+1}$ represents the student model at time $t+1$. This method, which does not employ selective updating or preservation of parameters, is likely to lead to a decline in model performance. Thus, in situations of continual domain shifts, it is crucial to retain necessary parameters to prevent catastrophic forgetting.

To retain the crucial parameters within the teacher model, we can use the diagonal FIM to determine whether this parameter is crucial. However, it is important to note that the input in this method is no longer the previously existing knowledge, but rather the batch of data $x_{t}$ at the present time $t$. The formulation of the diagonal FIM of the teacher model is as follows:
\begin{equation}\label{fim2}
    F^{'} = {\rm{Diag}}[\nabla\log p_{\theta_{t}}(x_{t})(\nabla\log p_{\theta_{t}}(x_{t}))^{T}].
\end{equation}
Subsequently, we can combine the aforementioned diagonal FIM to use the following formula to distinguish whether the parameter is crucial:
\begin{equation}\label{m}
    \textbf{m}_{j}=
\begin{cases}
1, ~~~~\text{if}~~~F_{j}^{'}<\text{quantile}(F^{'},\xi)\\
0, ~~~~\text{otherwise}
\end{cases},~j=1,2,\cdots,J,
\end{equation}
where quantile$(F^{'},\xi)$ \cite{quantile} is a threshold value which is the $\xi$-quantile of $F^{'}$, parameters with FIM value less than $\epsilon$ are set to 1. These parameters are considered significant and do not undergo a updating process. Conversely, parameters with FIM value equal to or greater than $\epsilon$ are set to 0 and updated according to Eq. \eqref{ema} using the EMA method. Specifically, the selective EMA mechanism can be described using the following formula:
\begin{equation}\label{teacher}
    \theta_{t+1}^{'}=\textbf{m} \odot \theta_{t}^{'} + (1-\textbf{m}) \odot (\delta\theta_{t}^{'}+(1-\delta)\theta_{t+1}).
\end{equation}
Here, $\odot$ denotes element-wise multiplication. This mechanism refines the traditional method of global parameter updating via an EMA by incorporating a restoration approach, thereby effectively mitigating the issues of excessive model drift and catastrophic forgetting.
\renewcommand{\algorithmicrequire}{\textbf{Input:}}
\renewcommand{\algorithmicprocedure}{\textbf{Procedure:}}
\renewcommand{\algorithmicensure}{\textbf{Output:}}
\begin{algorithm}[t]
  	\caption{The Proposed PSMT at time $t$}
  	\label{alg}
  	\begin{algorithmic}[1]
  		\Require Unlabeled test data $x_t$.
            \item[]\hspace*{-1.7em}\textbf{Init Model:} Pretrained model $f_{\theta_{0}}$, teacher model $f_{\theta^{'}_{0}}$.
            \item[]\hspace*{-1.7em}\textbf{Prediction:}
  		\State Augment input ${x}_{t}$ and generate pseudo-labels from the teacher model;
            \State Prediction from student model;
            \State Prediction $\hat{y}$ from teacher model;
            \State Compute consistency loss in Eq. \eqref{ce};
            \item[]\hspace*{-1.7em}\textbf{Adaptation:}
            \State Compute FIM in student model via Eq. \eqref{fim}; 
            \State Compute student loss by Fisher information in Eq. \eqref{stuloss};
            \State Update student model by \textbf{selective distillation} in Eq. \eqref{loss};
            \State Compute FIM in teacher model via Eq. \eqref{fim2}; 
            \State Compute mask \textbf{m} in Eq. \eqref{m};
  		\State Update teacher model by \textbf{selective EMA} via Eq. \eqref{teacher};
            \Ensure $\hat{y}$
  	\end{algorithmic}
\end{algorithm}
\subsection{Overall Update}
Following CoTTA \cite{CoTTA}, we employ a distillation loss:
\begin{equation}\label{ce}
    \mathcal{L}_{\text{ce}}=-\frac{1}{C}\sum\limits_{c=1}^{C}\hat{y}(c)\log{y}(c),
\end{equation}
where ${y}(c)$ represents the student logits, $\hat{y}(c)$ denotes the teacher logits, and $C$ signifies the number of categories.

All in all, in the ongoing process of model self-adaptation, the update is implemented using the loss formulated as follows. 
\begin{equation}\label{loss}
\mathcal{L}=\mathcal{L}_{\text{ce}}+\lambda\mathcal{L}_{\text{stu}}.
\end{equation}

As shown in \algoref{alg}, our entire process consists of learning regularized new knowledge and recovering important parameters, which together define our PSMT method. In this method, the data at time $t$ is predicted, followed by an adaptation using the predicted outcomes. Our proposed algorithm is articulated in lines 7 and 10 of \algoref{alg}. Line 7 corresponds to selective distillation, and line 10 executes selective EMA operations by selecting crucial parameters.

\section{Experiment}
\subsection{Experimental Setting}

\vspace{0.5em}
\noindent\textbf{4.1.1~~~Datasets}\\[0.5em]
We evaluate our method on CIFAR10C, CIFAR100C, and ImageNet-C \cite{corruption}, which are corrupted variants of the original CIFAR10 \cite{datasetcifar}, CIFAR100 \cite{datasetcifar}, and ImageNet \cite{datasetimagenet}, respectively. These modified datasets include real-world image corruptions to test resilience across different conditions. 
For the standard CIFAR10-to-CIFAR10C, CIFAR100-to-CIFAR100C, and ImageNet-to-ImageNet-C tasks, we employ a source model to adapt to fifteen different target domains with a corruption severity level of 5. Furthermore, in the case of gradually changing tasks, we apply the three datasets previously described, with corruption severity levels increasing from 1 to 5, then subsequently decreasing back to 1. In addition, for the ImageNet-C dataset, our methodology extends to performing analyses across ten different sequences, all at a corruption severity of 5.

\begin{table}[t]
	\caption{Classification error rate (\%) for the standard CIFAR10-to-CIFAR10C online continual test-time adaptation task.}
 \label{cifar10}
	\scriptsize
	\setlength{\tabcolsep}{0.8pt}
	\renewcommand{\arraystretch}{1.3}
	\centering
		\begin{tabular}{lcccccccccccccccc}
  \toprule
 \multirow{1}{*}{Time}
 & \multicolumn{15}{c}{$t \xrightarrow{\hspace{8.5cm}}$} \\
		\midrule
		\textbf{Method} & \textbf{GS} & \textbf{ST} & \textbf{IP} & \textbf{DF} &\textbf{GA} & \textbf{MO} & \textbf{ZM} & \textbf{SN} & \textbf{FT} & \textbf{FG} & \textbf{BT} & \textbf{CT} & \textbf{ES} & \textbf{PE} & \textbf{JG} & \textbf{Avg} \\ 
		\midrule
		Source \cite{source10} & 72.3 & 65.7 & 72.9 & 46.9 & 54.3 & 34.8 & 42.0 & 25.1 & 41.3 & 26.0 & 9.3 & 46.7 & 26.6 & 58.5 & 30.3& 43.5\\
			BN Adapt \cite{BN} & 28.1 & 26.1 & 36.3 &12.8&35.3&14.2&12.1&17.3&17.4&15.3&8.4&12.6&23.8&19.7&27.3&20.4\\
            SAR \cite{sar} & 28.3 & 26.0& 35.8&12.7&34.8&13.9&12.0&17.5&17.6&14.9&8.2&13.0&23.5&19.5&27.2&20.3\\
			TENT-cont \cite{tent} &24.8&20.6&28.5&15.1&31.7&16.9&15.6&18.3&18.3&18.1&11.0&16.8&23.9&18.6&23.9&20.1\\
            EATA \cite{eata} & 24.3& 19.1& 27.0& 12.4& 29.9& 13.9& 11.8& 16.5& 15.5& 15.0& 9.4& 12.5& 21.6& 16.8& 21.0& 17.8\\
			CoTTA \cite{CoTTA}   & 24.4 & 21.7 & 26.2 & 11.8 & 27.8 & 12.2 & 10.4 & 14.8 & 14.3 & 12.6 & 7.5 & 10.9 & 18.5 & 13.5 & 17.7 & 16.3 \\
			PETAL \cite{PETAL} & 23.4 & 21.1 & 25.7 & 11.7 & 27.2 & 12.2 & 10.3 & 14.8 & 13.9 & 12.7 & \textbf{\textcolor{red}{7.4}} & 10.5 & 18.1 & 13.4 & 16.8 & 15.9 \\
			LAW \cite{LAW} & 24.5 & 19.0 & 25.4 & 12.8 & 26.8 & 13.5 & 10.4 & 14.2 & 13.5 & 13.0 & 8.5 & 10.2 & 17.5 & 12.3 & 15.4 & 15.8 \\
			DSS \cite{DSS} & 24.1 & 21.3 & 25.4 & 11.7 & 26.9 & \textbf{\textcolor{red}{12.2}} & 10.5 & 14.5 & 14.1 & 12.5 & 7.8 & 10.8 & 18.0 & 13.1 & 17.3 & 16.0 \\
			Ours  & \textbf{\textcolor{red}{22.8}} & \textbf{\textcolor{red}{18.9}} & \textbf{\textcolor{red}{23.2}} & \textbf{\textcolor{red}{11.2}} & \textbf{\textcolor{red}{24.4}} & 12.3 & \textbf{\textcolor{red}{10.2}} & \textbf{\textcolor{red}{13.7}} & \textbf{\textcolor{red}{13.0}} & \textbf{\textcolor{red}{11.4}} & 7.8 & \textbf{\textcolor{red}{9.5}} & \textbf{\textcolor{red}{16.2}} & \textbf{\textcolor{red}{11.8}} & \textbf{\textcolor{red}{15.4}} &\textbf{\textcolor{red}{14.8}} \\
		\bottomrule
	\end{tabular}
\end{table}
\begin{table}[t]
	\caption{Classification error rate (\%) for the standard CIFAR100-to-CIFAR100C online continual test-time adaptation task.}
	\scriptsize
 \label{cifar100}
	\setlength{\tabcolsep}{0.8pt}
	\renewcommand{\arraystretch}{1.3}
	\centering
	\begin{tabular}{lcccccccccccccccc}
 \toprule
 \multirow{1}{*}{Time}
 & \multicolumn{15}{c}{$t \xrightarrow{\hspace{8.5cm}}$} \\
		\midrule
		\textbf{Method} & \textbf{GS} & \textbf{ST} & \textbf{IP} & \textbf{DF} &\textbf{GA} & \textbf{MO} & \textbf{ZM} & \textbf{SN} & \textbf{FT} & \textbf{FG} & \textbf{BT} & \textbf{CT} & \textbf{ES} & \textbf{PE} & \textbf{JG} & \textbf{Avg} \\ 
		\midrule
	Source \cite{source10} & 73.0 & 68.0 & 39.4 & 29.3 & 54.1 & 30.8 & 28.8 & 39.5 & 45.8 & 50.3 & 29.5 & 55.1 & 37.2 & 74.7 & 41.2& 46.4\\
			BN Adapt \cite{BN} & 42.1 & 40.7 & 42.7 &27.6&41.9&29.7&27.9&34.9&35.0&41.5&26.5&30.3&35.7&32.9&41.2&35.4\\
            SAR \cite{sar} & 40.5& 34.9& 37.1& 25.7& 37.2& 28.1& 25.6& 31.9& 30.9& 35.9& 25.1& 27.8& 31.8& 29.0& 37.3& 31.9\\
			TENT-cont \cite{tent} &37.2&35.8&41.7&37.9&51.2&48.3&48.5&58.4&63.7&71.1&70.4&82.3&88.0&88.5&90.4&60.9\\
            EATA \cite{eata} & \textbf{\textcolor{red}{37.2}}& 36.8& 37.4& 28.0& 37.6& 30.3& 27.3&32.6& 32.0& 35.7& 27.1& 29.2& 33.8& 29.6& 37.9& 32.8\\
			CoTTA \cite {CoTTA}   & 40.8 & 38.2 & 39.9 & 27.3 & 37.9 & 28.3 & 26.4& 33.5 & 32.2 & 40.2 & 25.0 & 26.9 & 32.4 & 28.3 & 33.9 & 32.7 \\
			PETAL \cite {PETAL}   & 38.3 & 36.4 & 38.6 & 25.9 & 36.8 & \textbf{\textcolor{red}{27.3}}& 25.4 & 32.0 & 30.8 & 38.7 & 24.4 & 26.4 & 31.5 & 26.9 & 32.5 & 31.5 \\
			LAW \cite{LAW} & 40.4 & 36.2 & 37.9 & \textbf{\textcolor{red}{25.8}} & 37.0 & 27.4 & 25.1 & 30.6 & 29.2 & 36.6 & 24.4 & 27.0 & 31.2 & 27.9 & 34.8 & 31.4 \\
			DSS \cite{DSS} & 39.7 & 36.0 & 37.2 & 26.3 & 35.6 & 27.5 & \textbf{\textcolor{red}{25.1}} & 31.4 & 30.0 & 37.8 & \textbf{\textcolor{red}{24.2}} & \textbf{\textcolor{red}{26.0}} & 30.0 & 26.3 & 31.1 & 30.9 \\
			Ours  & 39.6 & \textbf{\textcolor{red}{35.3}} & \textbf{\textcolor{red}{37.1}} & 26.2 & \textbf{\textcolor{red}{34.0}} & 27.8 & 25.4 & \textbf{\textcolor{red}{29.0}} & \textbf{\textcolor{red}{28.5}} & \textbf{\textcolor{red}{34.1}} & 24.7 & 26.1 & \textbf{\textcolor{red}{28.8}} & \textbf{\textcolor{red}{25.9}} & \textbf{\textcolor{red}{28.9}} & \textbf{\textcolor{red}{30.1}} \\
		\bottomrule
	\end{tabular}
\end{table}
\begin{table}[t]
	\caption{Classification error rate (\%) for the standard ImageNet-to-ImageNet-C online continual test-time adaptation task.}
 \label{imagenet}
	\scriptsize
	\setlength{\tabcolsep}{0.8pt}
	\renewcommand{\arraystretch}{1.3}
	\centering
	\begin{tabular}{lcccccccccccccccc}
 \toprule
 \multirow{1}{*}{Time}
 & \multicolumn{15}{c}{$t \xrightarrow{\hspace{8.5cm}}$} \\
		\midrule
		\textbf{Method} & \textbf{GS} & \textbf{ST} & \textbf{IP} & \textbf{DF} &\textbf{GA} & \textbf{MO} & \textbf{ZM} & \textbf{SN} & \textbf{FT} & \textbf{FG} & \textbf{BT} & \textbf{CT} & \textbf{ES} & \textbf{PE} & \textbf{JG} & \textbf{Avg} \\ 
		\midrule
		Source \cite{sourceimage} & 95.3 & 94.6 & 95.3 & 84.9 & 91.1 & 86.8 & 77.2 & 84.4 & 80.0 & 77.3 & 44.4 & 95.6 & 85.2 & 76.9 & 66.7& 77.2\\
			 BN Adapt \cite{BN} & 87.6 & 87.4 & 87.8 &87.7&88.0&78.2&64.5&67.6&70.6&54.9&36.4&89.3&58.0&56.4&66.6&66.2\\
    SAR \cite{sar}& 82.0& 80.9& 81.1& 81.2& 81.0& 69.4& 57.8& 61.9& 65.5& 49.1& 34.4& 77.0& 53.5& 48.1& 55.7&65.2\\
			TENT-cont \cite{tent}&85.7&80.0&78.3&82.2&79.2&70.9&59.1&65.6&66.4&55.4&40.6&80.3&55.5&53.5&59.0&67.4\\
   EATA \cite{eata} & 82.4& 76.9& 73.9& 77.4& 73.1& 63.9& 54.0&60.9& 61.2& 49.1& 36.0& 67.3& 49.4& 45.6& 49.9& 61.4\\
			CoTTA \cite{CoTTA} & 84.6 & 82.0 & 80.8 & 81.1 & 78.9 & 68.6 & 57.9 & 60.5 & 60.9 & 47.8 & 35.7 & 65.8 & \textbf{\textcolor{red}{47.2}} & 41.0 & 45.5 & 62.6 \\
			PETAL \cite{PETAL}   & 87.4 & 85.8 & 84.1 & 84.8 & 83.6 & 72.9 & 62.2 & 64.0 & 63.3 & 51.6 & 40.2 & 72.5 & 51.4 & 46.0 & 50.8 & 66.7 \\
			LAW \cite{LAW} & 80.7 & \textbf{\textcolor{red}{73.7}} & \textbf{\textcolor{red}{70.9}} & 77.8 & \textbf{\textcolor{red}{73.8}} & \textbf{\textcolor{red}{64.0}} & 54.9 & 57.7 & 60.6 & 46.8 & \textbf{\textcolor{red}{36.4}}& 67.9 & 48.5 & 45.1 & 48.7 & 60.5 \\
			DSS \cite{DSS} & 84.6 & 80.4 & 78.7 & 83.9 & 79.8 & 74.9 & 62.9 & 62.8 & 62.9 & 49.7 & 37.4 & 71.0 & 49.5 & 42.9 & 48.2 & 64.6 \\
			Ours  & \textbf{\textcolor{red}{79.9}} & 76.8 & 74.4 & \textbf{\textcolor{red}{77.4}} & 76.8 & 65.6 & \textbf{\textcolor{red}{54.9}} & \textbf{\textcolor{red}{57.0}} & \textbf{\textcolor{red}{60.5}} & \textbf{\textcolor{red}{45.7}} & 36.5 & \textbf{\textcolor{red}{61.7}} & 47.9 & \textbf{\textcolor{red}{40.8}} & \textbf{\textcolor{red}{42.4}} & \textbf{\textcolor{red}{59.9}} \\
		\bottomrule
	\end{tabular}
\end{table}

\vspace{0.5em}
\noindent\textbf{4.1.2~~~Implementation Details}\\[0.5em]
 We set up our classification task experiments following CoTTA \cite{CoTTA}, and we conducted Fisher information visualization tasks to verify the effectiveness of our approach in retaining crucial parameters. In our experiments, CIFAR10C is mapped to WideResNet-28 \cite{source10}, CIFAR100C to ResNeXt-29 \cite{ResNeXt29}, and ImageNet to ResNet-50 \cite{sourceimage}. We follow the hyperparameter configuration as outlined in \cite{CoTTA}. The batch size is 200 for experiments. Additionally, for the weight balancing parameter, $\lambda$ is set to 500. 
\subsection{Major Results}

\vspace{0.5em}
\noindent\textbf{4.2.1~~~Comparison with State-of-the-art Methods}\\[0.5em]
The results for the CIFAR10-to-CIFAR10C, CIFAR100-to-CIFAR100C and Ima-\\geNet-to-ImageNet-C datasets are respectively shown in \tableref{cifar10}, \tableref{cifar100} and \tableref{imagenet}. In this table, we abbreviate gaussian, shot, impulse, defocus, glass, motion, zoom, snow, frost, fog, brightness, contrast, elastic, pixelate and jpeg as GS, ST, IP, DF, GA, MO, ZM, SN, FT, FG, BT, CT, ES, PE, JG respectively. 

From these three tables, we can make several observations. Firstly, it is noted that some methods perform commendably on the CIFAR100C dataset in \tableref{cifar100}, such as PETAL and EATA. PETAL's superior performance is attributed to its utilization of Fisher information to restore knowledge from the pretrained model, which rendered it more effective than CoTTA. However, this method simply utilizes Fisher information. EATA~\cite{eata} also employs Fisher regularization to mitigate the issue of forgetting, but EATA does not incorporate Fisher information to the mean teacher architecture, which has been verfied effectiveness in many unsupervised learning. Secondly, our proposed method has surpassed the state-of-the-art across multiple domains in these three datasets. For instance, in the CIFAR10C dataset, our error rate is recorded at 14.8\%, compared to the current state-of-the-art, which stands at 15.8\%. This improvement is due to the effective prevention of key parameter updates during the model's adaptive process, facilitated by our Fisher information in the student model and teacher model, thereby validating the efficacy of our proposed method. Lastly, it is also observed that our method exhibits certain limitations. For example, in the motion and brightness domains across the three datasets, our performance is slightly inferior to that of the state-of-the-art, a deficiency still likely caused by error accumulation.

\begin{figure}[t]
  \centering
  \includegraphics[width=0.48\textwidth]{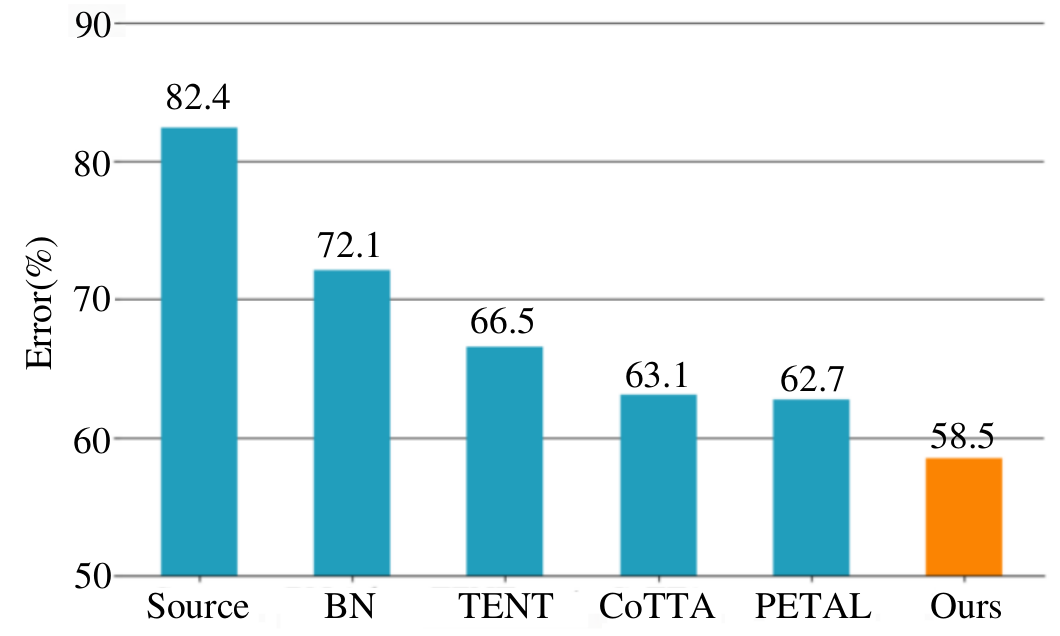}
  \caption{Average error rate on ImageNet-to-ImageNet-C for all corruption types over 10 varied sequences.}
  \label{imagenetfig}
\end{figure}

\vspace{0.5em}
\noindent\textbf{4.2.2~~~10 Different Orders of CTTA Loops}\\[0.5em]
Furthermore, to more effectively validate the robustness of our method and align our evaluation with CoTTA, we also conduct experiments on 10 different types of corruption sequences at corruption level 5, as depicted in \figref{imagenetfig}. Our method reduces the average classification error rate to 58.5\%, marking a 4.2\% improvement over PETAL. This outcomes further demonstrate better performance of our method in CTTA tasks.

\vspace{0.5em}
\noindent\textbf{4.2.3~~~Forgetting Evaluation with 3 Rounds}\\[0.5em]
We conduct experiments on five specific scenarios (fog, snow, frost, brightness, and pixelate) at corruption level 5 across three rounds. The purpose of the experiment is to verify whether our method can effectively mitigate the challenge of catastrophic forgetting. 
As shown in \tableref{round}, several observations can be made. 
First, our method demonstrates a reduction in error rates for each scenario after every learning round, achieving an average error rate of 11.6\%. This indicates that our approach can mitigate the phenomenon of catastrophic forgetting. Second, we also observe that in the brightness scenario, our method performs worse than the state-of-the-art in the first and second rounds. We attribute this to the possible occurrence of error accumulation.

\begin{table}[t]
	\caption{Classification error rate (\%) for three round of the CIFAR10-to-CIFAR10C online continual test-time adaptation task.}
 \label{round}
	\scriptsize
	\setlength{\tabcolsep}{0.8pt}
	\renewcommand{\arraystretch}{1.3}
	\centering
	\begin{tabular}{l|ccccc|ccccc|ccccc|c}
 \toprule
  \multirow{1}{*}{Round}
 & \multicolumn{5}{c|}{1}& \multicolumn{5}{c|}{2}& \multicolumn{5}{c|}{3}\\
		\midrule
		\textbf{Method} & \textbf{FG} & \textbf{SN} & \textbf{FT} & \textbf{BT} &\textbf{PE} & \textbf{FG} & \textbf{SN} & \textbf{FT} & \textbf{BT} &\textbf{PE} & \textbf{FG} & \textbf{SN} & \textbf{FT} & \textbf{BT} &\textbf{PE} & \textbf{Avg} \\ 
		\midrule
		Source \cite{sourceimage} & 26.0 & 25.1 & 41.3 & 9.3 & 58.5 & 26.0 & 25.1 & 41.3 & 9.3 & 58.5 & 26.0 & 25.1 & 41.3 & 9.3 & 58.5& 32.0\\
			 BN Adapt \cite{BN} & 14.9 & 17.5 & 17.6 &8.2&19.5&14.9&17.5&17.6&8.2&19.5&14.9&17.5&17.6&8.2&19.5.6&15.5\\
    SAR \cite{sar}& 14.9& 17.5& 17.6& 8.2& 19.5& 14.9& 17.5& 17.6& 8.2& 19.5& 14.9& 17.5& 17.6& 8.2& 19.5&15.5\\
			TENT-cont \cite{tent}&\textbf{\textcolor{red}{13.1}}&15.6&15.4&8.8&16.2&14.1&17.0&17.8&10.2&16.9&14.9&16.9&17.0&10.5&17.3&14.8\\
   EATA \cite{eata} & 13.2& 15.3& 15.0& 8.2& 15.0& 12.4& 14.3&14.4& 8.5& 14.9& 12.7& 14.3& 14.2& 8.9& 15.6& 13.1\\
			CoTTA \cite{CoTTA} & 14.6 & 15.4 & 15.4 & \textbf{\textcolor{red}{7.7}} & 15.6 & 13.7 & 14.7 & 14.9 & \textbf{\textcolor{red}{7.6}} & 14.9 & 13.7 & 14.6 & 14.5 & 7.6 & 14.5 & 13.3 \\
			PETAL \cite{PETAL}   & 14.1 & 15.3 & 15.1 & 7.6 & 15.3 & 13.8 & 14.7 & 14.8 & 7.6 & 15.0 & 13.6 & 14.3 & 14.1 & 7.4 & 14.8 & 13.2 \\
			LAW \cite{LAW} & 14.9 & 15.1 & 14.2 & 8.8 & 13.0 & 14.1 & 14.4 & 13.7 & 8.7 & 12.8 & 13.8 & 14.1 & 13.4 & 8.4 & 13.5 & 12.9 \\
			DSS \cite{DSS} & 13.2 & 15.3 & 15.2 & 8.0 & 13.8 & 12.9 & 14.9 & 14.8 & 8.0 & 13.6 & 12.6 & 14.7 & 14.4 & 7.7 & 13.1 & 12.8 \\
			Ours  & 13.4 & \textbf{\textcolor{red}{14.9}} & \textbf{\textcolor{red}{14.5}} & 7.9 & \textbf{\textcolor{red}{13.6}} & \textbf{\textcolor{red}{10.7}} & \textbf{\textcolor{red}{12.6}} & \textbf{\textcolor{red}{13.0}} & 7.7 & \textbf{\textcolor{red}{12.6}} & \textbf{\textcolor{red}{10.1}} & \textbf{\textcolor{red}{11.8}} & \textbf{\textcolor{red}{12.1}} & \textbf{\textcolor{red}{7.5}} & \textbf{\textcolor{red}{11.6}} & \textbf{\textcolor{red}{11.6}} \\
		\hline
	\end{tabular}
\end{table}

\subsection{Results on Gradual Test-Time Adaptation}
In addition to the aforementioned scenarios where the degree of corruption suddenly changes, we also conduct tests on the CIFAR10C, CIFAR100C, and ImageNet-C datasets for the task of gradual test-time adaptation \cite{GTTA} (GTTA). Unlike CTTA, GTTA features a gradual transition in the level of corruption across each type of corruption, as detailed below: 
\begin{equation*}
   \underbrace{\cdots\rightarrow 2 \rightarrow 1}_{\text{Domain}~t-1}
\xrightarrow[\text{change}]{\text{domain}} 
\underbrace{1 \rightarrow 2 \rightarrow \cdots \rightarrow 5\rightarrow\cdots1}_{\text{Domain}~t, \text{gradually changing severity}}
\xrightarrow[\text{change}]{\text{domain}}
\underbrace{1 \rightarrow 2\rightarrow3\cdots}_{\text{Domain}~t+1~\text{and on}}. 
\end{equation*}
\tableref{gradual} illustrates that our method achieves the lowest average classification error rates across these three datasets, recording 9.1\% for CIFAR10C, 25.4\% for CIFAR100C, and 40.0\% for ImageNet-C. Our approach yields enhancements of 1\%, 1.6\%, and 0.5\% over the LAW method for these datasets, respectively.
\begin{table}[t]
	\caption{Online mean classification errors (\%) for gradually changing CIFAR10-to-CIFAR10C, CIFAR100-to-CIFAR100C, and ImageNet-to-ImageNet-C.}\label{gradual}
	\setlength{\tabcolsep}{3.0pt}
	\centering
 \resizebox{.95\linewidth}{!}{
	\begin{tabular}{ccccccc}
		\toprule
		{} & \textbf{Source}\cite{sourceimage} & \textbf{BN Adapt}~\cite{BN} & \textbf{TENT-cont}\cite{tent} & \textbf{CoTTA}~\cite{CoTTA} & \textbf{LAW}~\cite{LAW}&\textbf{Ours}\\
		\hline
		\textbf{CIFAR10C} & 24.7 & 13.7 & 20.5 & 10.9 & 10.1 & \textbf{9.1} \\
		\textbf{CIFAR100C} & 33.6 & 29.9 & 74.8 & 27.0 & 27.0& \textbf{25.4} \\
		\textbf{ImageNet-C} & 58.4 & 48.3 & 46.4 & 43.6 & 40.5 & \textbf{40.0} \\
		\bottomrule
	\end{tabular}}
\end{table}
\subsection{Ablation Study and Analysis}

\vspace{0.5em}
\noindent\textbf{4.4.1~~~Effect of Each Module}\\[0.5em]
We conduct a series of ablation experiments on CIFAR10C, CIFAR100C, and ImageNet-C to verify the contributions of different modules within our method. In \tableref{ablation}, it is observable that upon incorporating the selective distillation module, the average error rates for the three datasets decreased by 1.3\%, 2.1\%, and 1.9\% respectively. This indicates that the inclusion of this module enables the model to rapidly assimilate new knowledge and mitigate error accumulation. In the second set of experiments, we remove the selective distillation module to elucidate the effect of solely incorporating the selective EMA module. It is found that, although the reduction in error rates is slightly less compared to the previous module, it still contributes to alleviating catastrophic forgetting by reinforcing existing knowledge. Upon integrating both modules into our methodology, the average error classification rates for the three scenarios decrease by 1.5\%, 2.6\%, and 2.7\%, respectively. These results further corroborate the efficacy of retaining crucial parameters in addressing error accumulation and catastrophic forgetting issues in the CTTA tasks.

\begin{table}[t]
	\caption{Ablation Study on the contribution of different modules. SD, SEMA represent the selective distillation and selective EMA, respectively.}\label{ablation}
	\setlength{\tabcolsep}{3.0pt}
	\centering
 \resizebox{.7\linewidth}{!}{
	\begin{tabular}{cccccc}
		\toprule
		\textbf{No.}&\textbf{SD} & \textbf{SEMA} & \textbf{CIFAR10C} & \textbf{CIFAR100C} & \textbf{ImageNet-C}\\
		\midrule
		\makecell{1}&{-} & {-}  & 16.3 & 32.7 & 62.6 \\
			\makecell{2}&{\checkmark} & {-}  & 15.0 & 30.6 & 60.7 \\
			\makecell{3}&{-} & {\checkmark}& 15.8 & 31.0 & 61.1 \\
			\makecell{4}&{\checkmark} & {\checkmark} & \textbf{14.8} & \textbf{30.1} & \textbf{59.9} \\
		\bottomrule
	\end{tabular}
 }
\end{table}

\vspace{0.5em}
\noindent\textbf{4.4.2~~~Hyperparameter Analysis}\\[0.5em]
Additionally, we conduct experiments to assess the hyperparameter $\lambda$ that balances the loss function in Eq. \eqref{loss}. As shown in \tableref{lam}, the optimal performance of the model on both the CIFAR10C and CIFAR100C datasets is achieved when $\lambda$ is set to 500. 
Concurrently, we also evaluate the hyperparameter $\xi$ that segments crucial parameters in Eq. \eqref{m}. As shown in \tableref{xi}, our method achieves the best performance when $\xi$ is set to 0.03.

\begin{table}[t]
    \centering
    \begin{minipage}{.45\linewidth}
      \centering
        \caption{Mean classification error (\%) with varing $\lambda$.}\label{lam}
        \resizebox{.8\linewidth}{!}{
        \begin{tabular}{ccc}
            \toprule            \boldmath{$\lambda$}&\textbf{CIFAR10C}&\textbf{CIFAR100C}\\
            \midrule
            {$50$}&{15.3}&30.8\\
            {$100$}&{15.2}&30.5\\
            {$500$}&\textbf{14.8}&\textbf{30.1}\\
            {$1000$}&{14.9}&30.9\\
            \bottomrule
        \end{tabular}}
    \end{minipage}%
    \hspace{10pt}
    \begin{minipage}{.45\linewidth}
      \centering
        \caption{Mean classification error (\%) with varing $\xi$.}\label{xi}
        \resizebox{.8\linewidth}{!}{
        \begin{tabular}{ccc}
            \toprule
            \boldmath{$\xi$}&\textbf{CIFAR10C}&\textbf{CIFAR100C}\\
            \midrule
            {$0.01$}&{15.1}& 30.3\\
            {$0.03$}&\textbf{14.8}&\textbf{30.1}\\
            {$0.05$}&{15.0}& 30.7\\
            {$0.1$}&{15.3}&30.8 \\
            \bottomrule
        \end{tabular}}
    \end{minipage}
\end{table}

\vspace{0.5em}
\noindent\textbf{4.4.3~~~Efficiency Evaluation}\\[0.5em]
To evaluate the efficiency of our methods, we leverage three metrics: time (ms), memory (GB), and FLOPs (GB). 
As shown in \tableref{efficency}, we evaluate the three metrics on one iteration and make several observations.
First, the efficiency of our method is comparable with some SOTA methods. The time consumption for predicting is 160.03 ms, which is significantly faster than the PEATL method, but slightly slower than CoTTA.
Second, we find that our method has higher memory usage. This is attributed to the storage of the FIM. Third, although our method does not exhibit the best performance across all efficiency metrics, it ensures robust performance without occupying excessive resources. We will continue to optimize the efficiency of our method in the future.

\begin{table}[h]
	\caption{Evaluation of methods' efficiency.}
 \label{efficency}
	\setlength{\tabcolsep}{3pt}
	\centering
 \resizebox{.6\linewidth}{!}{
		\begin{tabular}{lccc}
		\toprule
		\textbf{Method} & \textbf{Time(ms)} & \textbf{Memory(GB)}&  \textbf{FLOPs(GB)}\\ 
		\midrule
		CoTTA \cite{CoTTA} & 151.93 & 1.23&357.06\\
	   PEATL \cite{PETAL} &240.18 & 7.54&591.03\\
    Ours & 160.03 & 5.35&420.09\\
		\bottomrule
	\end{tabular}}
\end{table}

\begin{figure}[t]
	\includegraphics[width=\textwidth]{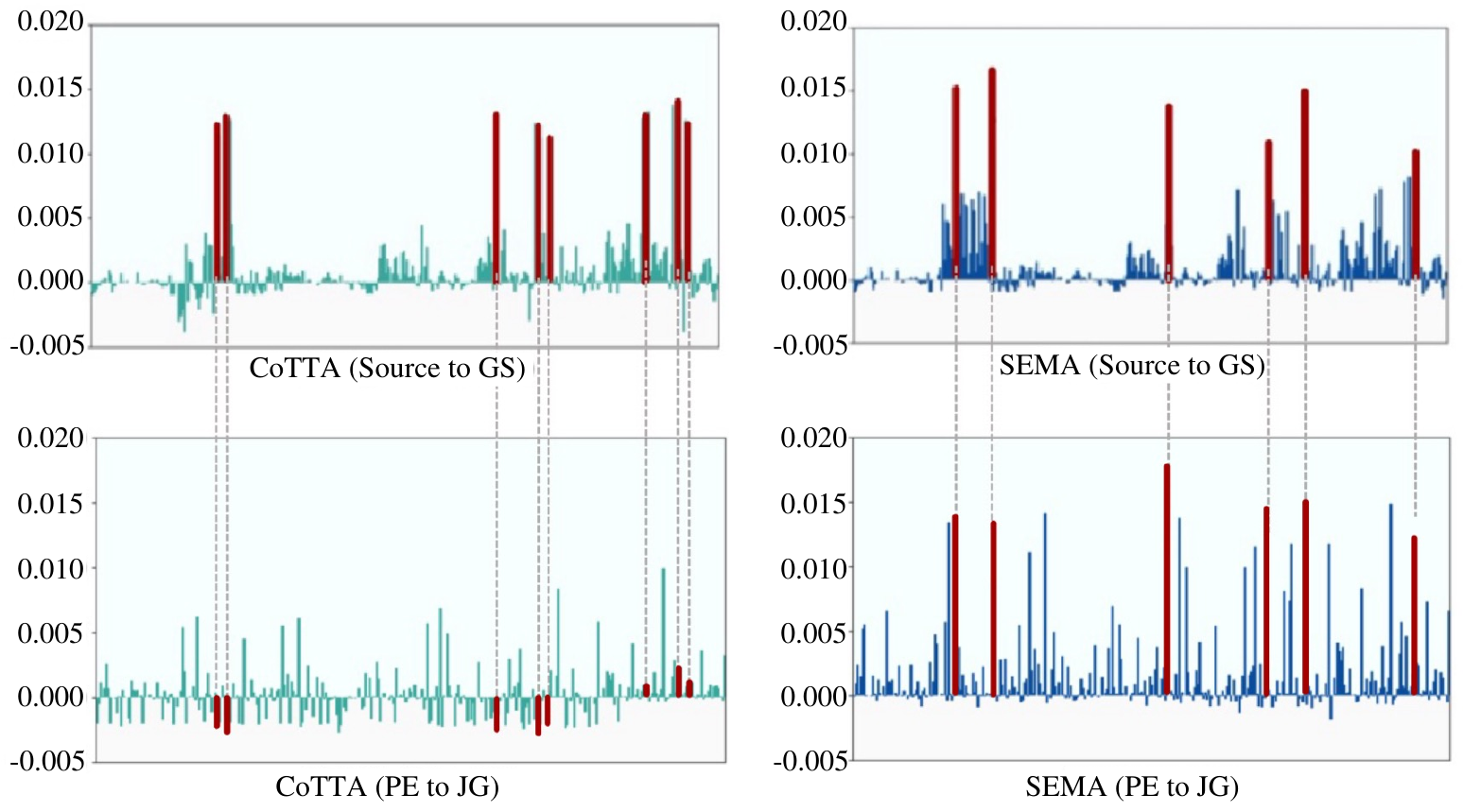}
	\caption{The performance of Fisher information in CoTTA method and selective EMA is assessed during domain shifts on the CIFAR10-to-CIFAR10C dataset. SEMA
    represents the selective EMA.} \label{fisherview}
\end{figure}

\vspace{0.5em}
\noindent\textbf{4.4.4~~~Fisher Information Visualization}\\[0.5em]
To better highlight the effects of Fisher information within our method, \figref{fisherview} illustrates the variations in Fisher information during domain transitions from source to GS and PE to JG within the CIFAR10C dataset. By utilizing the Fisher information of the subsequent domain minus that of the preceding domain, we assess the effectiveness of our restoration of crucial parameters. Higher values indicate that the parameter is significant and should be preserved. Moreover, it can be observed that the initial high values identified by CoTTA are not maintained during subsequent domain shifts. In contrast, selective EMA effectively preserves these important parameters, ensuring that the initial high values are retained even in the presence of domain shifts. The results demonstrate that our proposed approach effectively restores crucial parameters to prevent updating them in the CTTA tasks, this also proves the efficacy of our method.

\section{Conclusion}
In this paper, we studied the CTTA tasks and addressed the drawback of the traditional MT method, which may harm crucial parameters. 
We proposed a novel approach PSMT which consists of two modules. Firstly, 
we applied regularization based on previously acquired knowledge to retain important prior knowledge, thereby mitigating the phenomenon of error accumulation.
Secondly, 
we created a mask using FIM, based on which we selectively updated parameters through EMA to preserve important parameters, thereby alleviating the phenomenon of catastrophic forgetting. 
Moreover, the effectiveness of our method was validated on extensive experiments. 

Furthermore, we identified several limitations in our method. First, the noise issues in the Fisher visualization, which we thought may be caused by error accumulation due to CTTA across multiple domains. Additionally, the efficiency of our method is slightly inferior compared to certain previous approaches. We thought this reduction in efficiency may be due to the computational overhead required for storing and utilizing FIM. We intend to address these issues in our future work to refine and enhance the method.


%
%
\bibliographystyle{splncs04}
\bibliography{main}

\begin{thebibliography}{10}
\providecommand{\url}[1]{\texttt{#1}}
\providecommand{\urlprefix}{URL }
\providecommand{\doi}[1]{https://doi.org/#1}

\bibitem{grad}
Aburass, S., Dorgham, O.: Performance evaluation of swin vision transformer model using gradient accumulation optimization technique. In: FTC (2023)

\bibitem{driving2}
Barab{\'a}s, I., Todoru{\c{t}}, A., Cordo{\c{s}}, N., Molea, A.: Current challenges in autonomous driving. In: MSE (2017)

\bibitem{PETAL}
Brahma, D., Rai, P.: A probabilistic framework for lifelong test-time adaptation. In: CVPR (2023)

\bibitem{error}
Chen, C., Xie, W., Huang, W., Rong, Y., Ding, X., Huang, Y., Xu, T., Huang, J.: Progressive feature alignment for unsupervised domain adaptation. In: CVPR (2019)

\bibitem{medical1}
Chen, Z., Ye, Y., Lu, M., Pan, Y., Xia, Y.: Each test image deserves a specific prompt: Continual test-time adaptation for 2d medical image segmentation. arXiv preprint arXiv:2311.18363  (2023)

\bibitem{sensitivity}
Christopher~Frey, H., Patil, S.R.: Identification and review of sensitivity analysis methods. Risk Anal.  (2002)

\bibitem{datasetimagenet}
Deng, J., Dong, W., Socher, R., Li, L.J., Li, K., Fei-Fei, L.: Imagenet: A large-scale hierarchical image database. In: CVPR (2009)

\bibitem{rmt}
D{\"o}bler, M., Marsden, R.A., Yang, B.: Robust mean teacher for continual and gradual test-time adaptation. In: CVPR (2023)

\bibitem{datadriven}
Dong, B., Shen, Z.: Image restoration: a data-driven perspective. In: ICIAM (2015)

\bibitem{catastrophic2}
French, R.M.: Catastrophic forgetting in connectionist networks. Trends Cognit. Sci.  (1999)

\bibitem{overfit}
Hawkins, D.M.: The problem of overfitting. J. chem. inf. comput. sci.  (2004)

\bibitem{sourceimage}
He, K., Zhang, X., Ren, S., Sun, J.: Deep residual learning for image recognition. In: CVPR (2016)

\bibitem{corruption}
Hendrycks, D., Dietterich, T.: Benchmarking neural network robustness to common corruptions and perturbations. arXiv preprint arXiv:1903.12261  (2019)

\bibitem{neural}
Hong, Z., Yue, C.P.: Efficient-grad: Efficient training deep convolutional neural networks on edge devices with grad ient optimizations. TECS  (2022)

\bibitem{surveillance}
Iqbal, M.J., Iqbal, M.M., Ahmad, I., Alassafi, M.O., Alfakeeh, A.S., Alhomoud, A.: Real-time surveillance using deep learning. Secur. Commun. Netw.  (2021)

\bibitem{underfitting}
Jabbar, H., Khan, R.Z.: Methods to avoid over-fitting and under-fitting in supervised machine learning (comparative study). Computer Science, Communication and Instrumentation Devices  (2015)

\bibitem{fisheroptim}
Jung, Y., Lee, I.: Optimal design of experiments for optimization-based model calibration using fisher information matrix. Reliab. Eng. Syst. Saf.  (2021)

\bibitem{EWC}
Kirkpatrick, J., Pascanu, R., Rabinowitz, N., Veness, J., Desjardins, G., Rusu, A.A., Milan, K., Quan, J., Ramalho, T., Grabska-Barwinska, A., et~al.: Overcoming catastrophic forgetting in neural networks. PNAS  (2017)

\bibitem{datasetcifar}
Krizhevsky, A., Hinton, G., et~al.: Learning multiple layers of features from tiny images  (2009)

\bibitem{pseudo}
Kumar, V., Lal, R., Patil, H., Chakraborty, A.: Conmix for source-free single and multi-target domain adaptation. In: WACV (2023)

\bibitem{selftrain}
Liu, H., Wang, J., Long, M.: Cycle self-training for domain adaptation. In: NeurIPS (2021)

\bibitem{overparam}
Liu, S., Yin, L., Mocanu, D.C., Pechenizkiy, M.: Do we actually need dense over-parameterization? in-time over-parameterization in sparse training. In: ICML (2021)

\bibitem{liu}
Liu, X., Masana, M., Herranz, L., Van~de Weijer, J., Lopez, A.M., Bagdanov, A.D.: Rotate your networks: Better weight consolidation and less catastrophic forgetting. In: ICPR (2018)

\bibitem{domainshift1}
Luo, Y., Zheng, L., Guan, T., Yu, J., Yang, Y.: Taking a closer look at domain shift: Category-level adversaries for semantics consistent domain adaptation. In: CVPR (2019)

\bibitem{ly}
Ly, A., Marsman, M., Verhagen, J., Grasman, R.P., Wagenmakers, E.J.: A tutorial on fisher information. J. Math. Psychol.  (2017)

\bibitem{lyu2024variational}
Lyu, F., Du, K., Li, Y., Zhao, H., Zhang, Z., Liu, G., Wang, L.: Variational continual test-time adaptation. arXiv preprint arXiv:2402.08182  (2024)

\bibitem{lyu2023measuring}
Lyu, F., Sun, Q., Shang, F., Wan, L., Feng, W.: Measuring asymmetric gradient discrepancy in parallel continual learning. In: ICCV (2023)

\bibitem{lyu2021multi}
Lyu, F., Wang, S., Feng, W., Ye, Z., Hu, F., Wang, S.: Multi-domain multi-task rehearsal for lifelong learning. In: AAAI (2021)

\bibitem{ce}
Mao, A., Mohri, M., Zhong, Y.: Cross-entropy loss functions: Theoretical analysis and applications. In: ICML (2023)

\bibitem{GTTA}
Marsden, R.A., D{\"o}bler, M., Yang, B.: Gradual test-time adaptation by self-training and style transfer. arXiv preprint arXiv:2208.07736  (2022)

\bibitem{catastrophic}
McCloskey, M., Cohen, N.J.: Catastrophic interference in connectionist networks: The sequential learning problem. In: Psychol Learn Motiv (1989)

\bibitem{eata}
Niu, S., Wu, J., Zhang, Y., Chen, Y., Zheng, S., Zhao, P., Tan, M.: Efficient test-time model adaptation without forgetting. In: ICML (2022)

\bibitem{sar}
Niu, S., Wu, J., Zhang, Y., Wen, Z., Chen, Y., Zhao, P., Tan, M.: Towards stable test-time adaptation in dynamic wild world. arXiv preprint arXiv:2302.12400  (2023)

\bibitem{LAW}
Park, J., Kim, J., Kwon, H., Yoon, I., Sohn, K.: Layer-wise auto-weighting for non-stationary test-time adaptation. In: WACV (2024)

\bibitem{catastrophic3}
Ratcliff, R.: Connectionist models of recognition memory: constraints imposed by learning and forgetting functions. Psychol. Rev.  (1990)

\bibitem{BN}
Schneider, S., Rusak, E., Eck, L., Bringmann, O., Brendel, W., Bethge, M.: Improving robustness against common corruptions by covariate shift adaptation. In: NeurIPS (2020)

\bibitem{shi2024controllable}
Shi, Z., Lyu, F., Liu, Y., Shang, F., Hu, F., Feng, W., Zhang, Z., Wang, L.: Controllable continual test-time adaptation. arXiv preprint arXiv:2405.14602  (2024)

\bibitem{selftrain1}
Sinha, S., Gehler, P., Locatello, F., Schiele, B.: Test: Test-time self-training under distribution shift. In: WACV (2023)

\bibitem{domainshift}
Stacke, K., Eilertsen, G., Unger, J., Lundstr{\"o}m, C.: Measuring domain shift for deep learning in histopathology. IEEE J. Biomed  (2020)

\bibitem{tri}
Su, Y., Xu, X., Jia, K.: Towards real-world test-time adaptation: Tri-net self-training with balanced normalization. In: AAAI (2024)

\bibitem{nature}
Subburaj, J., Murugan, K., Keerthana, P., Aalam, S.S.: Catastropheguard: A guard against natural catastrophes through advances in ai and deep learning technologies. In: Internet of Things and AI for Natural Disaster Management and Prediction (2024)

\bibitem{tan2024less}
Tan, J., Lyu, F., Ni, C., Feng, T., Hu, F., Zhang, Z., Zhao, S., Wang, L.: Less is more: Pseudo-label filtering for continual test-time adaptation. arXiv preprint arXiv:2406.02609  (2024)

\bibitem{mean}
Tarvainen, A., Valpola, H.: Mean teachers are better role models: Weight-averaged consistency targets improve semi-supervised deep learning results. In: NeurIPS (2017)

\bibitem{quantile}
Tsong, C.C., Lee, C.F.: Quantile cointegration analysis of the fisher hypothesis. J MACROECON  (2013)

\bibitem{tent}
Wang, D., Shelhamer, E., Liu, S., Olshausen, B., Darrell, T.: Tent: Fully test-time adaptation by entropy minimization. arXiv preprint arXiv:2006.10726  (2020)

\bibitem{CoTTA}
Wang, Q., Fink, O., Van~Gool, L., Dai, D.: Continual test-time domain adaptation. In: CVPR (2022)

\bibitem{fisherlabel}
Wang, Q., Shen, X., Wang, M., Boyer, K.L.: Label consistent fisher vectors for supervised feature aggregation. In: ICPR (2014)

\bibitem{DSS}
Wang, Y., Hong, J., Cheraghian, A., Rahman, S., Ahmedt-Aristizabal, D., Petersson, L., Harandi, M.: Continual test-time domain adaptation via dynamic sample selection. In: WACV (2024)

\bibitem{sce}
Wang, Y., Ma, X., Chen, Z., Luo, Y., Yi, J., Bailey, J.: Symmetric cross entropy for robust learning with noisy labels. In: ICCV (2019)

\bibitem{ResNeXt29}
Xie, S., Girshick, R., Doll{\'a}r, P., Tu, Z., He, K.: Aggregated residual transformations for deep neural networks. In: CVPR (2017)

\bibitem{ess}
Yang, X., Gu, Y., Wei, K., Deng, C.: Exploring safety supervision for continual test-time domain adaptation. In: IJCAI (2023)

\bibitem{driving}
Yurtsever, E., Lambert, J., Carballo, A., Takeda, K.: A survey of autonomous driving: Common practices and emerging technologies. IEEE Access  (2020)

\bibitem{source10}
Zagoruyko, S., Komodakis, N.: Wide residual networks. arXiv preprint arXiv:1605.07146  (2016)

\end{thebibliography}
\end{document}